\documentclass[runningheads]{llncs}
\usepackage{graphicx}
\usepackage{tikz}
\usepackage{comment}
\usepackage{amsmath,amssymb} 
\usepackage{color}
\usepackage{graphicx}
\usepackage{booktabs}
\usepackage{bm}
\usepackage{multirow}
\usepackage{makecell}

\usepackage[accsupp]{axessibility}  

\usepackage[colorlinks,linkcolor=blue,anchorcolor=blue]{hyperref}

\usepackage[width=122mm,left=12mm,paperwidth=146mm,height=193mm,top=12mm,paperheight=217mm]{geometry}

\begin{document}
\pagestyle{headings}
\mainmatter

\title{3D Siamese Transformer Network for Single Object Tracking on Point Clouds} 

\makeatletter
\newcommand{\printfnsymbol}[1]{%
	\textsuperscript{\@fnsymbol{#1}}%
}
\makeatother

\titlerunning{3D Siamese Transformer Network}
%
\author{Le Hui \and Lingpeng Wang \and Linghua Tang \and Kaihao Lan \and Jin Xie\thanks{Corresponding authors.} \and Jian Yang\printfnsymbol{1}}
\authorrunning{L.Hui, L.Wang, L.Tang, K.Lan, J.Xie, J.Yang}
%
\institute{Key Lab of Intelligent Perception and Systems for High-Dimensional Information of Ministry of Education \\ Jiangsu Key Lab of Image and Video Understanding for Social Security \\ PCA Lab, School of Computer Science and Engineering \\ Nanjing University of Science and Technology, China \\
\email{\{le.hui,cslpwang,tanglinghua,lkh,csjxie,csjyang\}@njust.edu.cn}}
\maketitle
 
\begin{abstract}
Siamese network based trackers formulate 3D single object tracking as cross-correlation learning between point features of a template and a search area. Due to the large appearance variation between the template and search area during tracking, how to learn the robust cross correlation between them for identifying the potential target in the search area is still a challenging problem. In this paper, we explicitly use Transformer to form a 3D Siamese Transformer network for learning robust cross correlation between the template and the search area of point clouds. Specifically, we develop a Siamese point Transformer network to learn shape context information of the target. Its encoder uses self-attention to capture non-local information of point clouds to characterize the shape information of the object, and the decoder utilizes cross-attention to upsample discriminative point features. After that, we develop an iterative coarse-to-fine correlation network to learn the robust cross correlation between the template and the search area. It formulates the cross-feature augmentation to associate the template with the potential target in the search area via cross attention. To further enhance the potential target, it employs the ego-feature augmentation that applies self-attention to the local $k$-NN graph of the feature space to aggregate target features. Experiments on the KITTI, nuScenes, and Waymo datasets show that our method achieves state-of-the-art performance on the 3D single object tracking task. Source code is available at \url{https://github.com/fpthink/STNet}.
\keywords{3D Single Object Tracking, Siamese Network, Transformer, Point Clouds}
\end{abstract}

\section{Introduction}
Object tracking is a classic task in computer vision, and contributes to various applications, such as autonomous driving~\cite{liu2020deep,luo2018fast,lee2015road}, visual surveillance~\cite{xing2010multiple,tang2017multiple}, robotics vision~\cite{liu2016formation,comport2004robust}. Early efforts~\cite{kristan2015visual,kristan2016novel,bertinetto2016fully,danelljan2017eco,valmadre2017end} focus on visual object tracking that uses RGB images obtained by cameras. Recently, with the development of 3D sensors, such as LiDAR, 3D data is easy to acquire and set up 3D object tracking. Single object tracking is an important task in 3D computer vision. For example, it can improve the safety of autonomous vehicles by predicting the trajectory of key targets. However, due to the sparsity and irregular distribution of 3D points, existing popular schemes on 2D visual tracking cannot be directly applied to 3D single object tracking. Therefore, how to effectively track 3D objects in the complex scene is still a challenging problem.

Recently, Siamese network based trackers have raised much attention in the 3D single object tracking task. In~\cite{giancola2019leveraging}, Giancola~\emph{et al.} first proposed a shape completion based 3D Siamese tracker, which encodes shape information into a template to improve the matching accuracy between the template and candidate proposals in the search area. However, it is time-consuming and not an end-to-end method. To this end, Qi~\emph{et al.}~\cite{Qi2020P2BPN} proposed the point-to-box (P2B) network, which can be trained end-to-end and has a shorter inference time. Based on PointNet++~\cite{qi2017pointnet++}, P2B employs a target-specific feature augmentation module for the cross-correlation operation and adopts VoteNet~\cite{qi2019deep} to regress the target center from the search area. Based on P2B, zheng~\emph{et al.}~\cite{zheng2021box} proposed a box-aware tracker by inferring the size and the part priors of the target object from the template to capture the structure information of the target. Shan~\emph{et al.}~\cite{shan2021ptt} added a self-attention module in the VoteNet. Due to sparse point clouds, VoteNet is not suitable for regressing the target center in outdoor scenes. Lately, based on the bird's eye view feature map, Cui~\emph{et al.}~\cite{cui2021lttr} used cross-attention to learn the 2D relationship between the template and search to localize the target. In addition, Hui~\emph{et al.}~\cite{hui2019v2b} proposed a voxel-to-BEV tracker, which regresses the target center from the dense BEV feature map after performing shape completion in the search area. Nonetheless, due to the large appearance variation between template and search area, these simple cross-correlation feature learning cannot effectively characterize the correlation between them well.

In this paper, we propose a novel 3D Siamese Transformer tracking framework, which explicitly uses Transformer to learn the robust cross correlation between the template and search area in 3D single object tracking. Specifically, we first develop a Siamese point Transformer network by learning long-range contextual information of point clouds to extract discriminative point features for the template and search area, respectively. Our Siamese point Transformer network is an encoder-decoder structure. On each layer of the encoder, after aggregating the local features of the point cloud, we develop a non-local feature embedding module, which uses self-attention to capture the non-local information of point clouds. It is desired that the points can utilize the non-local features from the same instance to capture the whole structure of the object, $i.e.$, shape information. In the decoder, we propose an adaptive feature interpolation module to propagate features from subsampled points to the original points to generate discriminative point features. Compared with the commonly used distance-based interpolation~\cite{qi2017pointnet++}, the adaptive feature interpolation can effectively obtain discriminative point features through the learnable weights of the attention. Once we obtain discriminative point features of the template and the search area, we further develop an iterative coarse-to-fine correlation network to learn the cross-correlation between them for localizing the target in the search area. It consists of a cross-feature augmentation module and an ego-feature augmentation module. In the cross-feature augmentation module, we fuse the two feature maps from the template and search area by building cross-correlation between them via cross-attention. In this way, the template information is embedded into the search area for localizing the potential target. Once we localize the potential target, we use ego-feature augmentation to further enhance the potential target by applying self-attention to the local $k$-NN graph in the feature space instead of using the common self-attention over the whole point clouds. By applying self-attention to the $k$-NN graph in the feature space, the point features with similar semantic information can be aggregated, thereby enhancing the potential target. By iteratively performing the cross-feature and ego-feature modules, we can obtain a more discriminative feature fusion map for identifying the target from the search area. Finally, we integrate the Siamese point Transformer network, the iterative cross-correlation network, and the detection network~\cite{hui2019v2b} to form the Siamese Transformer tracking framework. Experiments on the KITTI~\cite{Geiger2012AreWR}, nuScenes~\cite{nuscenes2019}, and Waymo~\cite{sun2020scalability} datasets demonstrate the effectiveness of the proposed method on the 3D single object tracking task.

The contributions of this paper are as follows:
\begin{itemize}
	\item We present a Siamese point Transformer network that explicitly uses the attention mechanism to form an encoder-decoder structure to learn the shape context information of the target.
	\item We develop an iterative coarse-to-fine correlation network that iteratively applies the attention mechanism to the template and the search area for learning robust cross-correlation between them.
	\item The proposed 3D Siamese Transformer network achieves state-of-the-art performance on the KITTI, nuScenes, and Waymo datasets in 3D single object tracking.
\end{itemize}

\section{Related Work}
\noindent\textbf{3D single object tracking.} Early single object tracking approaches~\cite{bolme2010visual,henriques2012exploiting} focus on 2D images. Recently, Siamese network based trackers~\cite{held2016learning,tao2016siamese,wang2017dcfnet,guo2017learning,zhu2018distractor} have significantly improved tracking performance compared to the traditional correlation filtering based trackers~\cite{henriques2014high,danelljan2014adaptive,danelljan2015learning,zhang2015joint}. However, due to the lack of accurate depth information in RGB images, visual object tracking may not be able to accurately estimate the depth to the target.

Previous methods~\cite{spinello2010layered,luber2011people,pieropan2015robust,bibi20163d} adopt RGB-D data for 3D single object tracking. RGB-D based trackers~\cite{liu2018context,kart2018make,Kart2019ObjectTB} heavily rely on RGB information and adopt the same schemes used in visual object tracking with additional depth information. Although these approaches can produce very promising results, they may fail when critical RGB information is degraded. Recently, researchers~\cite{giancola2019leveraging,pang2021model} have focused on using 3D point clouds for single object tracking. Giancola~\emph{et al.}~\cite{giancola2019leveraging} first proposed a shape completion based 3D Siamese tracker (SC3D) for single object tracking. It performs template matching between the template and plenty of candidate proposals generated by Kalman filter~\cite{gordon2004beyond} in the search area, where a shape completion network is applied to the template for capturing the shape information of the object. Based on SC3D, Feng~\emph{et al.}~\cite{Feng2020ANO} proposed a two-stage framework called Re-Track to re-track the lost objects of the coarse stage in the fine stage. However, SC3D cannot be end-to-end trained. To achieve end-to-end training, point-to-box (P2B)~\cite{Qi2020P2BPN} first localizes the potential target center in the search area and then generates candidate proposals for verification. Due to incomplete targets in point clouds, box-aware tracker~\cite{zheng2021box} based on P2B proposes a box-aware feature fusion module to embed the bounding box information given in the first frame to enhance the object features, where the size and part information of the template are encoded. Shan~\emph{et al.}~\cite{shan2021ptt} improved P2B by adding a self-attention module in the detector VoteNet~\cite{qi2019deep} to generate refined attention features for increasing tracking accuracy. In addition, Cui~\emph{et al.}~\cite{cui2021lttr} proposed a Transformer-based method that first uses 3D sparse convolution to extract features to form a 2D BEV feature map and then uses Transformer to learn the 2D relationship between the template and search to localize the target. Lately, to handle sparse point clouds, Hui~\emph{et al.}~\cite{hui2019v2b} proposed a Siamese voxel-to-BEV tracker, which contains a Siamese shape-ware feature learning network and a voxel-to-BEV target localization network. It performs shape generation in the search area by generating a dense point cloud to capture the shape information of the target.

\noindent\textbf{3D multi-object tracking.} Different from 3D single object tracking, 3D multi-object tracking (3D MOT) usually adopts the tracking-by-detection paradigm~\cite{wu20213d,shenoi2020jrmot,kim2021eagermot,scheidegger2018mono}. 3D MOT trackers usually first use 3D detectors~\cite{shi2019pointrcnn,shi2020points,qi2019deep} to detect object instances for each frame and then associate the detected objects across all frames. Early 3D multi-object tracking approaches~\cite{chiu2020probabilistic,weng20203d} adopt 3D Kalman filters to predict the state of associated trajectories and objects instances. In~\cite{weng20203d}, Weng~\emph{et al.} first used PointRCNN~\cite{shi2019pointrcnn} to obtain 3D detections from a LiDAR point cloud, and then combined 3D Kalman filter and Hungarian algorithm for state estimation and data association. With the wide adoption of deep neural networks, recent methods~\cite{zhang2019robust,weng2020joint,wang2020joint} use neural networks to learn the 3D appearance and motion features for increasing accuracy. Lately, Weng~\emph{et al.}~\cite{weng2020gnn3dmot} proposed a graph neural network that uses a graph neural network for feature interaction by simultaneously considering the 2D and 3D spaces. Yin~\emph{et al.}~\cite{yin2021center} first proposed CenterPoint to detect 3D objects on the point clouds and then used a greedy closest-point matching algorithm to associate objects frame by frame.

\noindent\textbf{Transformer and attention.} Transformer is first introduced in~\cite{vaswani2017attention}, which uses a self-attention mechanism~\cite{lin2017structured} to capture long-range dependences of language sequences. Based on the Transformer, some further improvements have been proposed in various sequential tasks, including natural language processing~\cite{devlin2018bert,dai2019transformer,yang2019xlnet}, speech processing~\cite{synnaeve2019end,luscher2019rwth}. Recently, Dosovitskiy~\emph{et al.}~\cite{dosovitskiy2020image} first proposed a vision Transformer for image recognition, which introduces a Transformer to handle non-sequential problems. The key idea is to split an image into patches, and feed the sequence of linear embeddings of these patches into a Transformer. After that, the Transformer is extended to various visual tasks, such as semantic segmentation~\cite{liu2021swin,wang2021pyramid}, object detection~\cite{carion2020end,zhu2020deformable}, object tracking~\cite{chen2021transformer}. Recently, Liu~\emph{et al.}~\cite{liu2021swin} proposed a hierarchical Transformer based on shifted windows to greatly reduce the computational cost while maintaining the capability to capture long-range dependencies in the data. For point cloud processing, Zhao~\emph{et al.}~\cite{zhao2020point} first proposed a point Transformer that applies the self-attention mechanism on the local neighborhood of point clouds to extract local features for 3D semantic segmentation. Lately, inspired by point Transformer, different 3D vision tasks apply Transformer to yield good performance, such as point cloud classification~\cite{guo2020pct}, point cloud based place recognition~\cite{hui2021pyramid}, 3D object detection~\cite{pan20213d,mao2021voxel}, 3D object tracking~\cite{cui2021lttr,shan2021ptt}, and 3D action recognition~\cite{fan2021point}.

\begin{figure*}[t]
	\centering
	\includegraphics[width=0.98\linewidth]{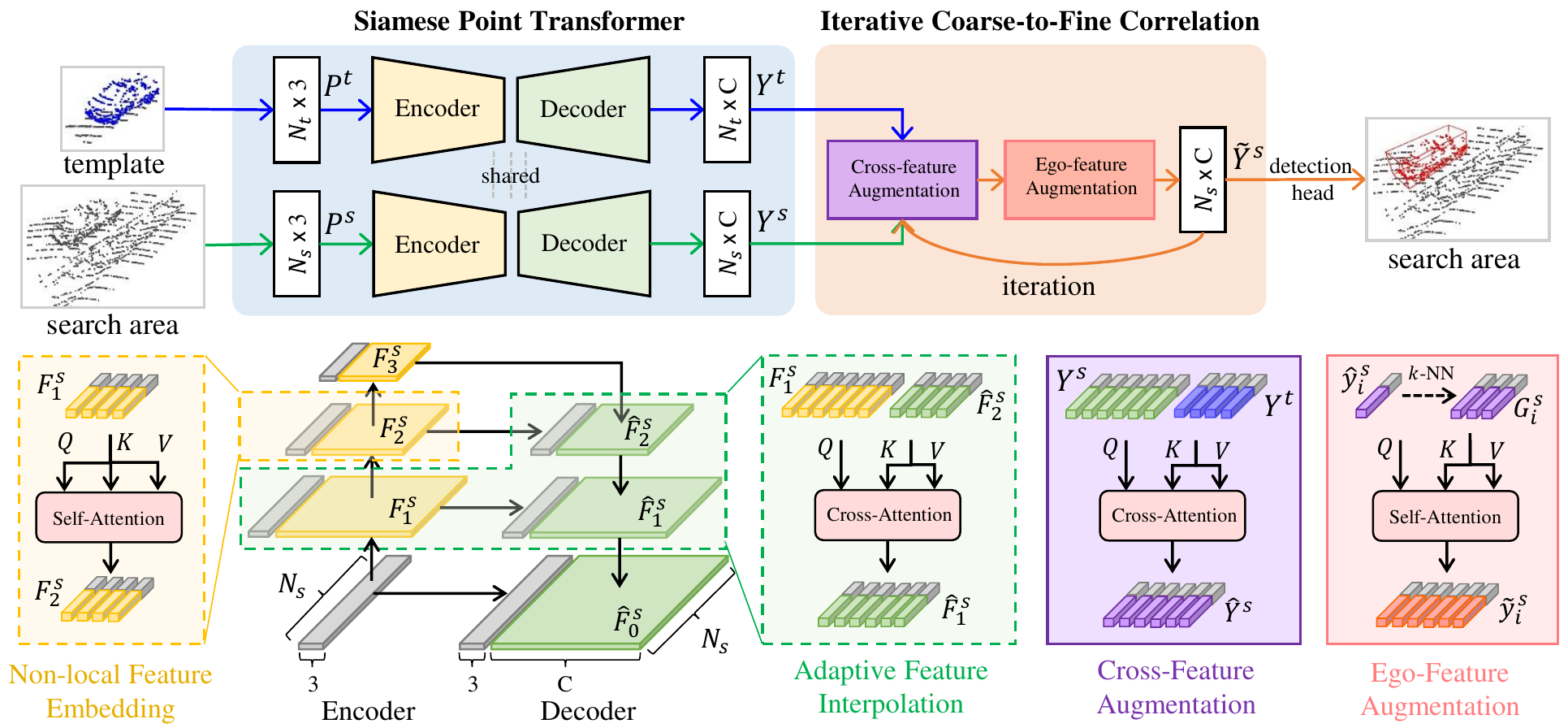}
	\caption{The framework of our Siamese Transformer network. Given the template $\bm{P}^t$ and search area $\bm{P}^s$, we first use the Siamese point Transformer network to extract features $\bm{Y}^t$ and $\bm{Y}^s$ for the template and search area, respectively. Then, we perform the iterative coarse-to-fine correlation network to obtain a feature fusion map $\widetilde{\bm{Y}}^s$. Finally, we apply the detection head on the feature fusion map to localize the target. Note that $Q$, $K$, and $V$ denote query, key, and value in Transformer, respectively.}
	\label{fig:framework}
\end{figure*}

\section{Method}\label{sec:method}

\subsection{Siamese Point Transformer Network}
In 3D single object tracking, given the target ($i.e.$, template) $\bm{P}^t=\{\bm{p}_i^t\}_{i=1}^{N_t}$ in the first frame, it aims to localize the 3D bounding box (3D BBox) of the same target in the search area $\bm{P}^s=\{\bm{p}_i^s\}_{i=1}^{N_s}$ frame by frame. $N_t$ and $N_s$ denote the number of points in the template and search area, and $\bm{p}_i^t$ and $\bm{p}_i^s$ are 3D coordinates. With a slight abuse of notations we use the same symbols for the sets of points and for their corresponding matrices $\bm{P}^t\in\mathbb{R}^{N_t\times 3}$ and $\bm{P}^s\in\mathbb{R}^{N_s\times 3}$. The 3D BBox is formed as a 7-dimensional vector, which contains box center ($x, y, z$), box size ($w, l, h$), and yaw angle $\theta$. Since the 3D BBox of the target is given in the first frame, we only need to regress the target center and yaw angle in the subsequent frames. By applying the displacement and yaw angle on the 3D BBox in the previous frame, the 3D BBox of the target in the current frame can be localized.

Most of existing Siamese trackers use local descriptors (such as PointNet~\cite{qi2017pointnet} and PointNet++~\cite{qi2017pointnet++}) as the feature extraction network. However, it lacks the ability to learn discriminative features by capturing long-range contextual information of point clouds. Thus, we propose a Siamese point Transformer network by utilizing attention to generate discriminative point features. As shown in Fig.~\ref{fig:framework}, it is a hierarchical feature learning network, consisting of two key modules: non-local feature embedding and adaptive feature interpolation.

\textbf{Non-local feature embedding.} The encoder consists of three non-local feature embedding modules. The non-local feature embedding module executes self-attention on feature maps at different scales, capturing the contextual information at different scales of the point cloud, respectively. Given the search area $\bm{P}^s$ of $N_s$ points, we follow P2B~\cite{Qi2020P2BPN} to downsample the point cloud to generate point clouds at different scales by using random sampling. The number of the subsampled points in the $l$-th layer is $\frac{N_s}{2^l}$.

Specifically, in the $l$-th layer, we first execute the local feature embedding to capture local geometric structures of point clouds. Inspired by~\cite{wang2018dynamic}, we apply edge convolution on the $k$-nearest neighbors ($k$-NN) in the coordinate space to aggregate local features, denoted by $\bm{E}_l^s\in\mathbb{R}^{\frac{N_s}{2^l}\times C_l}$. Then, we perform the self-attention on the feature map $\bm{E}_l^s$ to learn long-range context information of the point cloud. Formally, the attention mechanism is defined as:
\begin{equation}
\bm{F}_l^s = \operatorname{SelfAttention}(\bm{E}_l^s+\bm{X}_l^s, \bm{E}_l^s+\bm{X}_l^s, \bm{E}_l^s+\bm{X}_l^s)
\label{eqn:nonlocal}
\end{equation}
where $\bm{X}_l^s\in\mathbb{R}^{\frac{N_s}{2^l}\times C_l}$ denotes the position embedding of the sampled points in the $l$-th layer. Note that position information of the point cloud is very important, and thus we add the positional embedding to all matrices. In Eq.~(\ref{eqn:nonlocal}), the three inputs from left to right are used as query, key, and value, respectively. The obtained feature map $\bm{F}_l^s$ in the $l$-th layer will be used as the input of the ($l$+1)-th layer. In this way, we can obtain feature maps $\bm{F}_1^s, \bm{F}_2^s$, and $\bm{F}_3^s$ at three scales.

\textbf{Adaptive feature interpolation.} After the encoder, the original point set is subsampled. As the number of points on the target is reduced, it is difficult to identify the target accurately. Although the distance based interpolation~\cite{qi2017pointnet++} can be used to interpolate new points, it cannot effectively interpolate high-quality point features for the target, especially in sparse point clouds. Thus, we design a learnable interpolation module to interpolate point features from the subsampled points to the original points through the learnable weights of the attention.

We define $\bm{F}_0\in\mathbb{R}^{N_s\times 3}$ ($i.e.$, 3D coordinates) as the point feature of the original point with a size of $N_s$. Given the obtained feature maps $\bm{F}_1^s\in\mathbb{R}^{\frac{N_s}{2}\times C_1}, \bm{F}_2^s\in\mathbb{R}^{\frac{N_s}{4}\times C_2}$, $\bm{F}_3^s\in\mathbb{R}^{\frac{N_s}{8}\times C_3}$, and $\widehat{\bm{F}}_3^s$=$\bm{F}_3^s$, we gradually execute the adaptive feature interpolation to generate new point features from the low-resolution point cloud to the high-resolution point cloud, which is written as:
\begin{equation}
\widehat{\bm{F}}_l^s=\operatorname{CrossAttention}(\bm{F}_l^s, \widehat{\bm{F}}_{l+1}^s, \widehat{\bm{F}}_{l+1}^s + \bm{X}_{l+1}^s)
\label{eqn:adaptive}
\end{equation}
where $l\in\{2,1,0\}$ and $\widehat{\bm{F}}_l^s\in\mathbb{R}^{\frac{N_s}{2^l}\times C_l}$ is the interpolated feature map. $\bm{X}_{l+1}^s\in\mathbb{R}^{\frac{N_s}{2^l}\times C_l}$ is the positional embedding. Note that the query $\bm{F}_l^s$ is the high-resolution feature map features, while the key ($\widehat{\bm{F}}_{l+1}^s$) and value ($\widehat{\bm{F}}_{l+1}^s$+$\bm{X}_{l+1}^s$) are the low-resolution feature maps. In Eq.~(\ref{eqn:adaptive}), the feature map $\widehat{\bm{F}}_l^s$ is interpolated by weighting point features of the low-resolution (value) point cloud, considering the similarity between the high-resolution (query) and the low-resolution (key) point clouds. Finally, by applying the Siamese point Transformer on the template and search area, we obtain the feature maps $\widehat{\bm{F}}_0^t\in\mathbb{R}^{N_t\times C_0}$ and $\widehat{\bm{F}}_0^s\in\mathbb{R}^{N_s\times C_0}$ for the original point sets of the template and search area. For simplicity, we denote the obtained feature maps of the template and search area by $\bm{Y}^t\in\mathbb{R}^{N_t\times C}$ and $\bm{Y}^s\in\mathbb{R}^{N_s\times C}$. Note that $\bm{Y}^t=\widehat{\bm{F}}_0^t$ and $\bm{Y}^s=\widehat{\bm{F}}_0^s$.

\subsection{Iterative Coarse-to-Fine Correlation Network}
In 3D Siamese trackers, a cross-correlation operation is used to compute the similarity between the template and search area to generate a feature fusion map for identifying the target. Most of existing trackers use the cosine distance to generate the similarity map. Due to the large appearance variation between template and search area, this simple operation cannot effectively associate the template with the search area. Thus, we develop an iterative coarse-to-fine correlation network to learn the similarity in a coarse-to-fine manner to mitigate large appearance variation between them through the attention mechanism. Fig.~\ref{fig:framework} shows the detailed structure.

\textbf{Cross-feature augmentation.} We employ the cross-feature augmentation module to fuse the template and the search area by learning similarity between them. Given the template feature $\bm{Y}^{t}\in\mathbb{R}^{N_t\times C}$ and search area feature $\bm{Y}^{s}\in\mathbb{R}^{N_s\times C}$, we use the cross-attention mechanism between the template and search area to generate a coarse feature fusion map. Specifically, the cross-feature augmentation operation is formulated as:
\begin{equation}
\setlength{\abovedisplayskip}{5pt}
\setlength{\belowdisplayskip}{5pt}
\widehat{\bm{Y}}^s=\operatorname{CrossAttention}(\bm{Y}^s, \bm{Y}^t, \bm{Y}^t + \bm{X}^t)
\label{equ:ccc}
\end{equation}
where $\widehat{\bm{Y}}^s\in\mathbb{R}^{N_s\times C}$ is the obtained coarse feature fusion map. Since the 3D coordinates of the template provide the positional relationship of the target, we add the positional embedding of the template $\bm{X}^t\in\mathbb{R}^{N_s\times C}$ to the value $\bm{Y}^t\in\mathbb{R}^{N_s\times C}$. By learning the similarity between the template (key) and search area (query), we embed the template (value) into the search area to generate the feature fusion map $\widehat{\bm{Y}}^s$. In this way, the potential target in the feature fusion map can be associated with the template.

\textbf{Ego-feature augmentation.} Furthermore, we design an ego-feature augmentation module to enhance target information by considering the internal association in the coarse feature fusion map. Specifically, we first construct the $k$-nearest neighbor ($k$-NN) for each point in the feature space. Given the coarse feature fusion map $\widehat{\bm{Y}}^s\in\mathbb{R}^{N_s\times C}$ with $N_s$ feature vectors $\widehat{\bm{y}}_i^s\in\mathbb{R}^{C}$, the similarity between points $i$ and $j$ is denoted as:
\begin{equation}
\setlength{\abovedisplayskip}{5pt}
\setlength{\belowdisplayskip}{5pt}
a_{i,j}=\exp(-\|\widehat{\bm{y}}_i^s-\widehat{\bm{y}}_j^s\|_2^2)
\label{eqn:sim}
\end{equation}
where $a_{i,j}$ is the similarity metric. For the point $i$, we select $K$ nearest points in the feature space as its neighborhood by using the defined similarity. Thus, we obtain the local $k$-NN feature map for the point $i$, denoted by $\bm{G}_i^s\in\mathbb{R}^{K\times C}$. Since the points in the same instance have similar appearances, they are close to each other in the feature space. By aggregating local $k$-NN graphs in the feature space, the differences between different instances can be further magnified. Therefore, we then use the self-attention on the local $k$-NN graph to capture local association to aggregate discriminative point features. Specifically, given the point feature $\widehat{\bm{y}}_i^s\in\mathbb{R}^{C}$ and its $k$-NN feature map $\bm{G}_i^s\in\mathbb{R}^{K\times C}$, the local association is defined as:
\begin{equation}
\widetilde{\bm{y}}_i^s=\operatorname{SelfAttention}(\widehat{\bm{y}}_i^s+\bm{x}_i^s, \bm{G}_i^s+\bm{Z}_i^s, \bm{G}_i^s+\bm{Z}_i^s)
\label{equ:fsc}
\end{equation}
where $\bm{x}_i^s\in\mathbb{R}^C$ and $\bm{Z}_i^s\in\mathbb{R}^{K\times C}$ are the positional embeddings of the $i$-th point and its $k$-NN neighborhood, and $\widetilde{\bm{y}}_i^s\in\mathbb{R}^C$ is the extracted point feature. In this way, we obtain the refined feature fusion map $\widetilde{\bm{Y}}^s\in\mathbb{R}^{N_s\times C}$.

We iteratively perform the coarse cross-feature augmentation module and the fine ego-feature augmentation module to generate a discriminative feature fusion map for identifying the target. Note that the output in the previous iteration will replace the search area input in the next iteration. By capturing the external (template) and internal (search area itself) relationships, our iterative coarse-to-fine correlation network can gradually generate a discriminative feature fusion map for identifying the target. Based on the feature fusion map, we use the 3D detector~\cite{hui2019v2b} to regress the target center and yaw angle.

\section{Experiments}
\subsection{Experimental Settings}
\textbf{Datasets.} We use the KITTI~\cite{Geiger2012AreWR}, nuScenes~\cite{nuscenes2019}, and Waymo~\cite{sun2020scalability} datasets for single object tracking. For the KITTI dataset, it contains 21 video sequences. Following~\cite{giancola2019leveraging}, we split the sequences into three parts: sequences 0-16 for training, 17-18 for validation, and 19-20 for testing. For the nuScenes dataset, it contains 700 sequences for training and 150 sequences for validation. Since the ground truth for the test set in nuScenes is inaccessible offline, we use its validation set to evaluate our method. For the Waymo dataset, we follow LiDAR-SOT~\cite{pang2021model} to use 1,121 tracklets, which are split into easy, medium, and hard subsets according to the number of points in the first frame of each tracklet. Following~\cite{hui2019v2b}, we use the trained model on the KITTI dataset to test on the nuScenes and Waymo datasets for evaluating the generalization ability of our 3D tracker.

\textbf{Evaluation metrics.} We use \emph{Success} and \emph{Precision} defined in one pass evaluation~\cite{kristan2016novel} as the evaluation metrics for 3D single object tracking. Specifically, \emph{Success} measures the intersection over union (IOU) between the predicted 3D bounding box (BBox) and ground truth (GT) box, while \emph{Precision} measures the AUC for the distance between both two boxes' centers from 0 to 2 meters.

\textbf{Network architecture.} Following~\cite{Qi2020P2BPN}, we randomly sample $N_t=512$ for each template $P^t$ and $N_s=1024$ for each search area $P^s$. For the Siamese point Transformer network, it consists of a three-layer encoder (three non-local feature embedding modules) and a three-layer decoder (three adaptive feature interpolation modules). In each encoder layer, the number of points is reduced by half. For example, if we feed the search area of 1024 points to the encoder, the number of points in each layer is 512, 256, and 128, respectively. Besides, the neighborhood sizes used to extract local feature are 32, 48, and 48, respectively. In the decoder, it gradually aggregates feature maps layer by layer to obtain a discriminative feature map. The obtained feature map sizes of the template $\bm{Y}^t$ and search area $\bm{Y}^s$ are $512\times32$ and $1024\times 32$, respectively. For the iterative coarse-to-fine correlation network, we use two iterations considering the computational complexity and inference time. The hyperparameter $K$ of the neighborhood size is set to $48$. The size of the output feature fusion map $\widetilde{\bm{Y}}^s$ is $1024\times 32$. Note that in this paper, we adopt the linear Transformer~\cite{lin2020attn} and employ $n=2$ attention heads for all experiments.

\begin{table*}[t]
	\centering
	\caption{The performance of different methods on the KITTI and nuScenes datasets. Note that the results on the nuScenes dataset are obtained by using the pre-trained model on the KITTI dataset. ``Mean'' denotes the average results of four categories.}
	\resizebox{1.0\textwidth}{!}{
		\begin{tabular}{c|c|ccccc|ccccc}
			\toprule
			\multirow{3}*{} & Method &  &  & Success &  &  & & & Precision & &\\
			\hline
			\hline
			& Category & Car & Pedestrian & Van & Cyclist & Mean  & Car & Pedestrian & Van & Cyclist & Mean \\
			&Frame Num. & 6424 & 6088 & 1248 & 308 & 14068 & 6424 & 6088 & 1248 & 308 & 14068 \\
			\hline
			\multirow{8}*{\rotatebox{90}{KITTI}}&SC3D~\cite{giancola2019leveraging} & 41.3 & 18.2 & 40.4 & 41.5 & 31.2 & 57.9 & 37.8 & 47.0 & 70.4 & 48.5\\
			&P2B~\cite{Qi2020P2BPN} & 56.2 & 28.7 & 40.8 & 32.1 & 42.4 & 72.8 & 49.6 & 48.4 & 44.7 & 60.0\\
			&MLVSNet~\cite{wang2021mlvsnet} & 56.0 & 34.1 & 52.0 & 34.3 & 45.7 & 74.0 & 61.1 & 61.4 & 44.5 & 66.6\\
			&LTTR~\cite{cui2021lttr} & 65.0 & 33.2 & 35.8 & 66.2 & 48.7 & 77.1 & 56.8 & 45.6 & 89.9 & 65.8\\
			&BAT~\cite{zheng2021box} & 60.5 & 42.1 & 52.4 & 33.7 & 51.2 & 77.7 & 70.1 & 67.0 & 45.4 & 72.8\\
			&PTT~\cite{shan2021ptt} & 67.8 & 44.9 & 43.6 & 37.2 & 55.1  & 81.8 & 72.0 & 52.5 & 47.3 & 74.2\\
			&V2B~\cite{hui2019v2b} & 70.5 & 48.3 & 50.1 & 40.8 & 58.4  & 81.3 & 73.5 & 58.0 & 49.7 & 75.2\\
			&STNet (ours) & {\bf 72.1}  & {\bf 49.9}  & {\bf 58.0} & {\bf 73.5} & {\bf 61.3} & {\bf 84.0} & {\bf 77.2} & {\bf 70.6} & {\bf 93.7} & {\bf 80.1}\\
			\hline
			\hline
			& Category & Car & Pedestrian & Truck & Bicycle & Mean & Car & Pedestrian & Truck & Bicycle & Mean \\
			&Frame Num. & 15578 & 8019 & 3710 & 501 & 27808 & 15578 & 8019 & 3710 & 501 & 27808 \\
			\hline
			\multirow{5}*{\rotatebox{90}{nuScenes}}&SC3D~\cite{giancola2019leveraging}  & 25.0 & 14.2 & {\bf 25.7} & 17.0 & 21.8 & 27.1 & 16.2 & {\bf 21.9} & 18.2 & 23.1\\
			&P2B~\cite{Qi2020P2BPN}  & 27.0 & 15.9 & 21.5 & 20.0 & 22.9 & 29.2 & 22.0 & 16.2 & 26.4 & 25.3\\
			&BAT~\cite{zheng2021box}  & 22.5 & 17.3 & 19.3 & 17.0 & 20.5 & 24.1 & 24.5 & 15.8 & 18.8 & 23.0\\
			&V2B~\cite{hui2019v2b} & 31.3 & 17.3 & 21.7 & {\bf 22.2} & 25.8 & 35.1 & 23.4 & 16.7 & 19.1 & 29.0 \\
			&STNet (ours) & {\bf 32.2} & {\bf 19.1} &22.3 &21.2 & {\bf 26.9} & {\bf 36.1} & {\bf 27.2} & 16.8 & {\bf 29.2} & {\bf 30.8}\\
			\bottomrule
		\end{tabular}
	}
	\label{tab:all_results}
\end{table*}

\subsection{Results}
\begin{figure*}[t]
	\centering
	\includegraphics[width=0.98\linewidth]{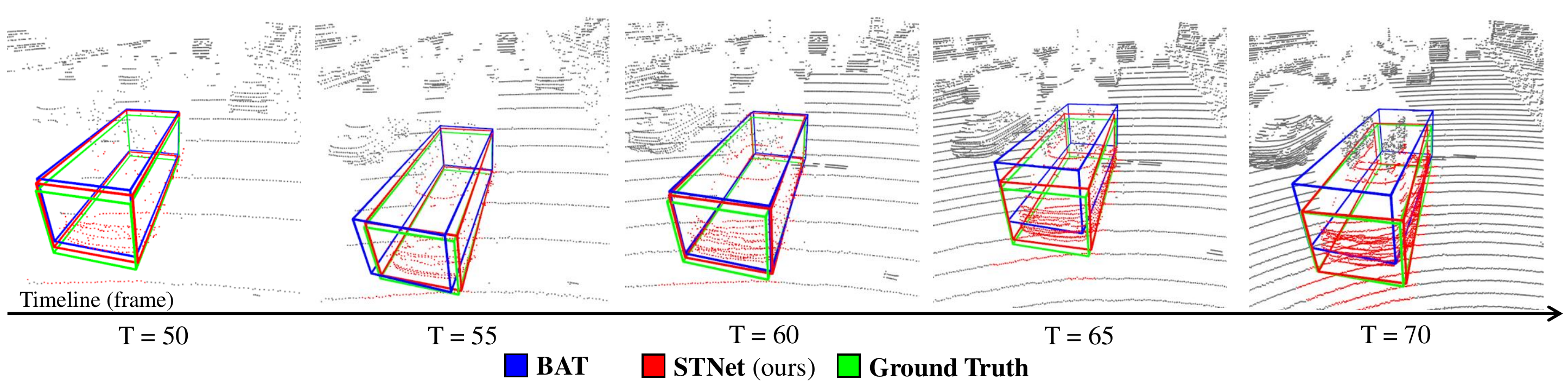}
	\caption{The visualization results of our STNet and BAT on the car category of the KITTI dataset. The points on the target car are colored in red.}
	\label{fig:vis}
\end{figure*}

\textbf{Quantitative results.} As shown in the top half of Tab.~\ref{tab:all_results}, we make comprehensive comparisons on the KITTI dataset with the previous state-of-the-art methods, including SC3D~\cite{giancola2019leveraging}, P2B~\cite{Qi2020P2BPN},
MLVSNet~\cite{wang2021mlvsnet},
BAT~\cite{zheng2021box}, LTTR~\cite{cui2021lttr}, PTT~\cite{shan2021ptt}, and V2B~\cite{hui2019v2b}. Following~\cite{Qi2020P2BPN}, we report the results over four categories, including car, pedestrian, van, and cyclist. From the table, it can be found that our method outperforms other methods on the mean results of four categories. For the car category, our method can significantly improve the precision from 81.8\% to 84.0\% with a gain of about 2\% on the car category. In addition to the large targets, our method can still achieve higher performance for those small targets, such as cyclists. For the cyclist category, compared with LTTR, our method obtains a gain of about 3\% on the precision. In Fig.~\ref{fig:vis}, we also show the visualization results of our method and BAT. It can be observed that our method (red boxes) can accurately localize the target. Most existing methods use local descriptors to extract point features. Due to the large appearance variation between template and search area, the extracted features cannot characterize the differences between them well. Thus, we propose a Siamese point Transformer network to learn the dense and discriminative point features with a learnable point interpolation module, where the shape context information of the target can be captured. Furthermore, we use an iterative coarse-to-fine correlation network to learn the similarity between the template and search area in a coarse-to-fine manner to mitigate large appearance variations in sparse point clouds for accurate object localization.

\begin{figure*}[t]
	\centering
	\includegraphics[width=0.98\linewidth]{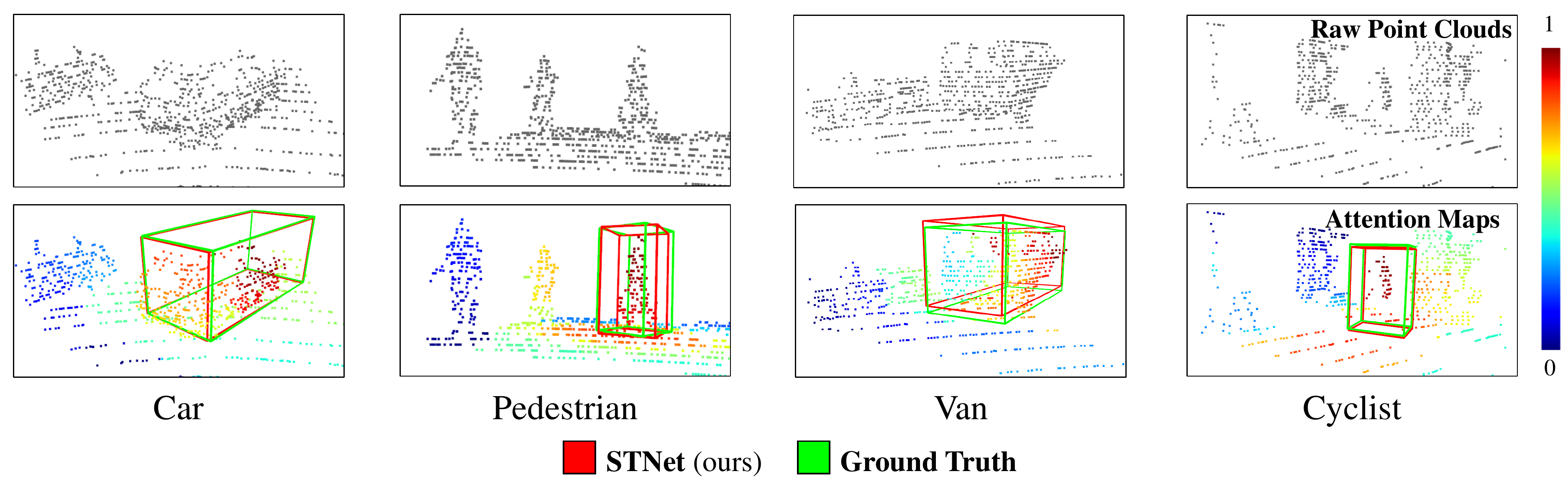}
	\caption{The attention maps generated by our method on the KITTI dataset.}
	\label{fig:attnmap2}
\end{figure*}

\textbf{Visualization of attention maps.} In Fig.~\ref{fig:attnmap2}, we show the attention maps generated by our method on the KITTI dataset, including car, pedestrian, van, and cyclist. The points marked with the red color can obtain high attentional weights. It can be observed that our method can accurately focus on the target in the search area. The visualization results show that when there are multiple objects, the target can be distinguished from the non-target objects. It means that the learned shape context information of the object can help to learn the discriminative relationship between the template and search area by Transformer.

\begin{table*}
	\centering
	\caption{The performance of different methods on the Waymo dataset. Each category is split into three levels of difficulty: ``Easy'', ``Medium'', and ``Hard''. ``Mean'' denotes the average results of three levels. Note that except for our STNet, the results of other methods are obtained by running the official codes.}
	\resizebox{1.0\textwidth}{!}{
		\begin{tabular}{c|c|cccc|cccc}
			\toprule
			& Method & \multicolumn{4}{c|}{Vehicle}& \multicolumn{4}{c}{Pedestrian}\\
			\hline
			\hline
			& Split & Easy & Medium & Hard & Mean & Easy & Medium & Hard & Mean \\
			& Frame Num. & \;67832\; & \;61252\; & \;56647\; & \;185731\; & \;85280\; & \;82253\; & \;74219\; & \;241752\; \\
			\hline
			\multirow{4}*{Success} & P2B~\cite{Qi2020P2BPN} & 57.1 & 52.0 & 47.9 & 52.6 & 18.1 & 17.8 & 17.7 & 17.9\\
			& BAT~\cite{zheng2021box} & 61.0 & 53.3 & 48.9 & 54.7 & 19.3 & 17.8 & 17.2 & 18.2\\
			& V2B~\cite{hui2019v2b} & 64.5 & 55.1 & 52.0 & 57.6 & 27.9 & 22.5 & 20.1 & 23.7\\
			& STNet (ours) & {\bf 65.9} & {\bf 57.5} & {\bf 54.6} & {\bf 59.7} & {\bf 29.2} & {\bf 24.7} & {\bf 22.2} & {\bf 25.5}\\ 
			\hline
			\multirow{4}*{Precision} & P2B~\cite{Qi2020P2BPN} & 65.4 & 60.7 & 58.5 & 61.7 & 30.8 & 30.0 & 29.3 & 30.1\\
			& BAT~\cite{zheng2021box} & 68.3 & 60.9 & 57.8 & 62.7 & 32.6 & 29.8 & 28.3 & 30.3\\
			& V2B~\cite{hui2019v2b} & 71.5 & 63.2 & 62.0 & 65.9 & 43.9 & 36.2 & 33.1 & 37.9\\
			& STNet (ours) & {\bf 72.7} & {\bf 66.0} & {\bf 64.7} & {\bf 68.0} & {\bf 45.3} & {\bf 38.2} & {\bf 35.8} & {\bf 39.9}\\ 
			\bottomrule
		\end{tabular}
	}
	\label{tab:waymo_results}
\end{table*}

\textbf{Generalization ability.} To verify the generalization ability of our method, we transfer the trained model of the KITTI dataset to obtain the testing results on the nuScenes and Waymo datasets. Following~\cite{hui2019v2b}, we use the pre-trained models on four categories (car, pedestrian, van, cyclist) of the KITTI dataset to evaluate the corresponding categories (car, pedestrian, truck, bicycle) on the nuScenes dataset. The results are listed in the bottom half of Tab.~\ref{tab:all_results}. Note that except for our results, the results of other methods are taken from paper~\cite{hui2019v2b}. It can be observed that our method outperforms other methods on the mean results of four categories. In addition, Tab.~\ref{tab:waymo_results} shows the results of vehicle and pedestrian categories on the Waymo dataset. It can be observed that our method outperforms other methods in terms of different subsets, including easy, medium, and hard. KITTI and Waymo datasets are built by 64-beam LiDAR, while nuScenes dataset is built by 32-beam LiDAR. Due to the large discrepancy between data distributions of datasets caused by different LiDAR sensors and sparsity of point clouds, it is very challenging to directly use the pre-trained model of the KITTI dataset to generalize it on nuScenes. The previous method V2B only achieves the performance of 25.8\%/29.0\% on the average of four categories. Our method achieves the gains of +1.1\%/+1.8\% over V2B. However, due to similar data distributions of the KITTI and Waymo datasets, the generalization results of the Waymo dataset are higher than those in the nuScenes dataset. The generalization results further demonstrate the effectiveness of our method for unseen scenes.

\textbf{Ability to handle sparse scenes.} We report the results of different methods in the sparse scenarios. Following~\cite{hui2019v2b}, we divide the number of points into four intervals, including [0, 150), [150, 1000), [1000, 2500), and [2500, $+\infty$). In Tab.~\ref{tab:interval}, we report average \emph{Success} and \emph{Precision} for each interval on the car category of the KITTI dataset. Note that except for our results, other results are taken from~\cite{hui2019v2b}. It can be observed that our method achieves the best performance on all four intervals. Especially in the sparse point clouds below 150 points, our method can improve the performance by about 2\% on both \emph{Success} and \emph{Precision} compared with V2B. Moreover, as the number of points increases, the performance of our method gradually increases. Due to the good results of our method in sparse point clouds, it will lead to more accurate template updates on the subsequent dense frames, resulting in better performance.

\textbf{Running speed.} We also report the average running time of all test frames in the car category on the KITTI dataset. Specifically, we evaluate our model on a single TITAN RTX GPU. Our method achieves 35 FPS, including 4.6 ms for processing point clouds, 22.7 ms for network forward propagation, and 1.3 ms for post-processing. In addition, on the same platform, V2B, P2B and SC3D in default settings run with 37 FPS, 46 FPS, and 2 FPS, respectively. Due to Transformer, the forward time of our method is longer than that in P2B. However, the performance of our method is significantly better than that of P2B.

\begin{table*}[t]
	\centering
	\caption{The average \emph{Success} and \emph{Precision} for each interval on the car category in the KITTI dataset.}
	\resizebox{1.0\textwidth}{!}{
		\begin{tabular}{c|cccc|cccc}
			\toprule
			 Method & \multicolumn{4}{c|}{Success} & \multicolumn{4}{c}{Precision} \\
			\hline
			\hline
			Intervals & {\small [0, 150)} & {\small [150, 1k)} & {\small [1k, 2.5k)} & {\small [2.5k, $+\infty$)} & {\small [0, 150)} & {\small [150, 1k)} & {\small [1k, 2.5k)} & {\small [2.5k, $+\infty$)}  \\
			Frame Num. & 3293 & 2156 & 693 & 282 & 3293 & 2156 & 693 & 282 \\
			\hline
			SC3D~\cite{giancola2019leveraging} & 37.9 & 36.1 & 33.8 & 23.7 & 53.0 & 53.1 & 48.7 & 35.3 \\
			P2B~\cite{Qi2020P2BPN} & 56.0 & 62.3 & 51.9 & 43.8 & 70.6 & 78.6 & 68.1 & 61.8 \\
			BAT~\cite{zheng2021box} & 60.7 & 71.8 & 69.1 & 61.6 & 75.5 & 83.9 & 81.0 & 72.9 \\
			V2B~\cite{hui2019v2b} & 64.7 & 77.5 & 72.3 & 82.2 & 77.4 & 87.1 & 81.5 & 90.1 \\
			STNet (ours) & {\bf 66.3}  & {\bf 77.9}  & {\bf 79.3} & {\bf 83.1}  & {\bf 79.9} & {\bf 87.8} & {\bf 89.6} & {\bf 91.0}  \\    
			\bottomrule
		\end{tabular}
	}
	\label{tab:interval}
\end{table*}

\subsection{Ablation Study}
In this section, we conduct the ablation study to validate the effectiveness of the designed modules. Due to a large number of test samples in the car category of the KITTI dataset, the ablated experiments on it can truly reflect the impact of different settings on the tracking accuracy.

\textbf{Non-local embedding and adaptive interpolation.} To verify the effectiveness of our Siamese point Transformer network, which adopts the non-local feature embedding and the adaptive feature interpolation, we conduct experiments to study their effects on performance. Specifically, we use PointNet++~\cite{qi2017pointnet++} as the baseline and add the non-local embedding (dubbed ``NL emb.'') and the adaptive feature interpolation (dubbed ``AF inte.'') to conduct experiments. The results are listed in Tab.~\ref{tab:ablation}. It can be observed that the performance of only PointNet++ is worse than using both non-local feature embedding and adaptive feature interpolation (``NL emb. + AF inte.''). According to the results in the ablation study, capturing shape context information of the object can effectively improve tracking performance.

\textbf{Coarse-to-fine correlation.} Here we conduct experiments to study the effects of coarse-to-fine correlation on performance. Specifically, we perform the simple feature augmentation used in P2B~\cite{Qi2020P2BPN}, only cross-feature augmentation (dubbed ``CF aug.''), and cross- and ego-feature augmentations (dubbed ``CF aug. + EF aug.''), respectively. The results are listed in Tab.~\ref{tab:ablation}. It can be found that when the cross- and ego-feature augmentations are used at the same time, we can obtain the best results. Since P2B's feature augmentation only uses the cosine distance to measure the similarity between the template and search area, it cannot obtain a high-quality feature fusion map. However, our method iteratively learns the similarity in a coarse-to-fine manner between them, so the target information in the feature fusion map can be further enhanced.

\begin{figure*}[t]
	\centering
	\includegraphics[width=1.0\linewidth]{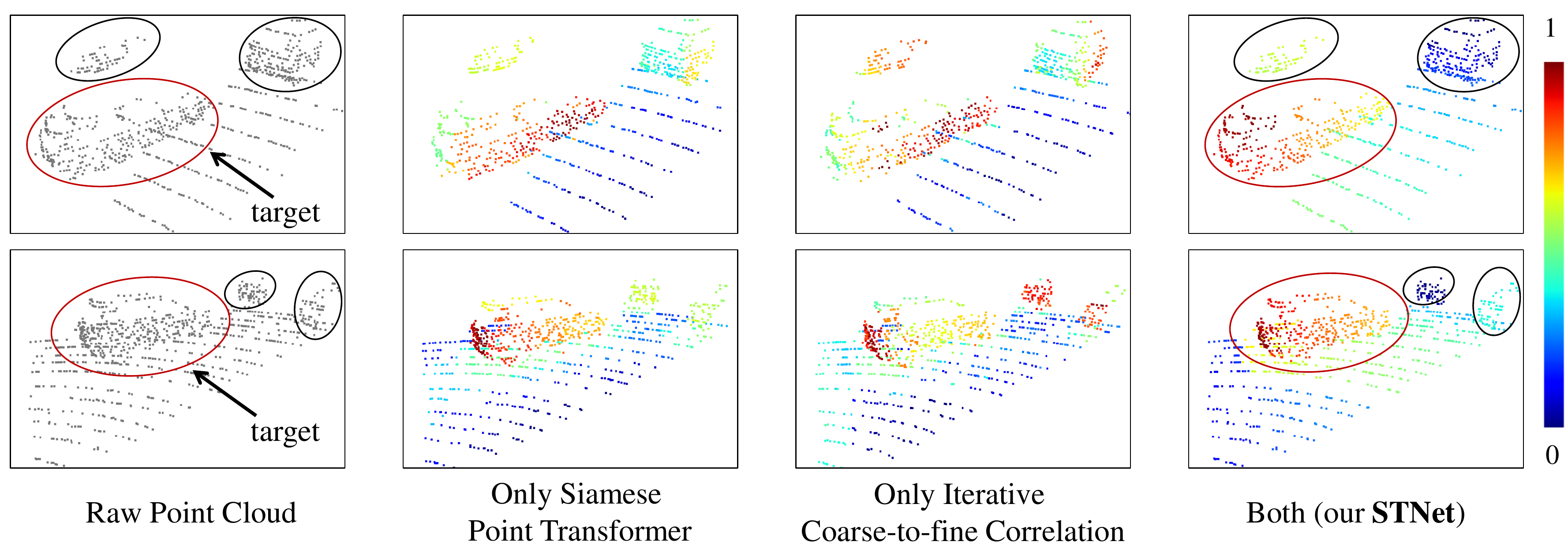}
	\caption{The attention maps generated by different components of our STNet on the KITTI dataset. The leftmost column shows the input raw point cloud, and we circle the target object to distinguish the background.}
	\label{fig:ablation3}
\end{figure*}

\textbf{Comparison of attention maps of different components.} In Fig.~\ref{fig:ablation3}, we show the attention maps of the extracted features using our proposed different components. It can be observed that only using the Siamese point Transformer or only using the iterative coarse-to-fine correlation cannot effectively focus on the target object, while STNet using all components is able to effectively distinguish the car from the background. Furthermore, it can be observed that the target car can be clearly recognized from three cars since learned shape information of the target is helpful to learn the discriminative relationship between the template and search area by Transformer.
\\
\makeatletter\def\@captype{table}\makeatother
\begin{minipage}{.55\textwidth}
	\centering
	\caption{The ablation study results of different components.}
	\resizebox{0.92\textwidth}{!}{
		\begin{tabular}{l|cc}
			\toprule
			\multicolumn{1}{c|}{Method} & \emph{Success} & \emph{Precision}\\
			\hline
			\hline
			PointNet++~\cite{qi2017pointnet++} & 66.1 & 76.9\\
			only NL emb. & 69.9 & 81.8\\
			only AF inte. & 68.1 & 80.9\\
			NL emb. + AF inte. & {\bf 72.1} & {\bf 84.0}\\
			\hline
			feature aug. in P2B~\cite{Qi2020P2BPN} & 69.4 & 80.8\\
			CF aug. & 71.0 & 82.4 \\
			CF aug. + EF aug. & {\bf 72.1} & {\bf 84.0}\\
			\bottomrule
		\end{tabular}
	}
	\label{tab:ablation}
\end{minipage}
\makeatletter\def\@captype{table}\makeatother
\begin{minipage}{.45\textwidth}
	\centering
	\caption{The ablation study results of different hyperparameters.}
	\resizebox{1.0\textwidth}{!}{
		\begin{tabular}{l|c|cc}
			\toprule
			\multicolumn{1}{c|}{STNet} & Parameters & \;\;\emph{Success}\;\; & \emph{Precision} \\
			\hline
			\hline
			\multirow{6}*{Neighbors} & $K=16$ & 68.1 & 79.6\\
			& $K=32$ & 71.0 & 82.4\\
			& $K=48$ & {\bf 72.1} & {\bf 84.0}\\
			& $K=64$ & 69.6 & 81.7 \\
			& $K=80$ & 68.7 & 80.5 \\
			& $K=96$ & 67.2 & 78.6 \\
			\hline
			\multirow{4}*{Iterations}& iter. $=$ 1 & 69.1 & 81.7\\
			&iter. $=$ 2 & {\bf 72.1} & 84.0\\
			&iter. $=$ 3 & 71.8 & {\bf 84.2}\\
			&iter. $=$ 4 & 72.0 & 84.0\\
			\bottomrule
		\end{tabular}
	}
	\label{tab:ablation2}
\end{minipage}

\textbf{Different $K$ in ego-feature augmentation.} The neighbor size $K$ is a key parameter in the ego-feature augmentation. Here we study the effects of different values of $K$ on tracking accuracy. In Tab.~\ref{tab:ablation2}, we report the performance in the cases of different neighbor sizes, including 16, 32, 48, 64, 80, and 96, respectively. It can be observed that when the neighbor size is set to 48, we can obtain the best results. If $K$ is too small, our transformer cannot characterize the local geometry structures of point clouds, while if $K$ is too large, background points in the scene are incorporated to localize the target.

\textbf{Different numbers of iterations.} We also study the effects of the number of iterations of coarse-to-fine correlation on performance. In Tab.~\ref{tab:ablation2}, we list the quantitative results in the cases of different numbers of iterations. It can be observed that the performance of our method with two iterations is comparable to that of our method with three or four iterations. As the number of iterations increases, the GPU memory will gradually increase. Thus, considering the memory consumption, we choose to iterate twice.

\section{Conclusions}
In this paper, we proposed a 3D Siamese Transformer framework for single object tracking on point clouds. We developed a Siamese point Transformer network that uses the attention mechanism to formulate a encoder-decoder structure to learn shape context information of the target. Also, we constructed an iterative coarse-to-fine correlation network to produce a feature fusion map by using the attention mechanism on the template and the search area. In this way, we can effectively associate the template and the search area in a coarse-to-fine manner so as to mitigate large appearance variations between them in sparse point clouds. The experiments show that the proposed method achieves state-of-the-art performance on the KITTI, nuScenes, and Waymo datasets on 3D single object tracking.

\clearpage
\bibliographystyle{splncs04}
\bibliography{egbib}
\end{document}